\pdfoutput=1

\documentclass[11pt]{article}

\usepackage[]{emnlp2021}

\usepackage{times}
\usepackage{latexsym}

\usepackage[T1]{fontenc}

\usepackage[utf8]{inputenc}

\usepackage{microtype}
\usepackage{tabularx}
\usepackage{booktabs}
\usepackage{graphicx}

\usepackage{lettrine}
\newcommand{\GrantNo}{825303}
\newcommand{\ProjectName}{Bergamot}
\newcommand{\ProjectType}{Research and Innovation Action}

\title{TranslateLocally: Blazing-fast translation running on the local CPU}

\author{Nikolay Bogoychev \and Jelmer Van der Linde \and Kenneth Heafield \\
        School of Informatics \\
        University of  Edinburgh \\
        \texttt{\{N.Bogoych,Jelmer.vanderLinde,Kenneth.Heafield\}@ed.ac.uk}}

\begin{document}
\maketitle
\begin{abstract}
Every day, millions of people sacrifice their privacy and browsing habits in exchange for online machine translation.  Companies and governments with confidentiality requirements often ban online translation or pay a premium to disable logging.  To bring control back to the end user and demonstrate speed, we developed \textit{translateLocally}.  Running locally on a desktop or laptop CPU, \textit{translateLocally} delivers cloud-like translation speed and quality even on 10 year old hardware.  The open-source software is based on Marian and runs on Linux, Windows, and macOS.  
\end{abstract}

\section{Introduction}

Neural Machine Translation \citep{bahdanau_nmt, vaswani_transformer} is pervasive but has a reputation for high computational cost. The combination of the typically high computational cost, however, has pushed its delivery to the cloud, with a number of cloud providers available (Google, Microsoft, Facebook, Amazon, Baidu, etc.). Using a cloud based translation provider carries an inherent privacy risk, as users lose control of their data once it enters the web. Potential issues include public disclosure due to not understanding terms of service \cite{translatecom}, contractors reading user data \cite{contractors}, use of user data for advertising, and data breaches.

To preserve privacy, we made a translation system that runs locally: \textit{translateLocally}.  Once a translation model is downloaded, it does not use an Internet connection.  Running locally is challenging due to a number of factors: the model needs to be small enough to download on a user hardware; translation latency can't be hidden by splitting and parallelising the translation of a large documents across multiple machines; consumer hardware has highly variable computing power; availability of GPU computational resources can't be assumed. 

We therefore focused on trimming model size and optimising speed for CPUs while aiming to preserve translation quality.  The result is fast enough that users see translations update as they type with latency comparable to ping times to the cloud.  

Targeting non-expert users, the open-source (primarily MIT) software\footnote{\url{https://github.com/XapaJIaMnu/translateLocally}} is also available as compiled binaries for Linux, Windows and Mac from the official webpage: \url{https://translatelocally.com}.  Translation models for several language pairs are provided, while advanced users can add their own models.  

\section{Design}
Our product is based on the Marian machine translation toolkit \cite{mariannmt}, heavily optimised for speed with a Qt based GUI.

\subsection{Translation Engine}
For the translation engine core, we used the same Marian fork as the one used by \citet{Bogoychev-wngt20} for participating in the 2020 Workshop on Neural Generation and Translation's efficiency shared task \citep[WNGT 2020,][]{heafield-etal-2020-findings}. We introduce binary lexical shortlists and streamlined binary model loading to the codebase, resulting in a comparable translation speed, but slightly faster loading time. We also add sentence splitting and formatting preservation are handled by a C++ wrapper around Marian.\footnote{\url{https://github.com/browsermt/bergamot-translator}} 

\subsection{Translation Models}

\begin{table*}[!ht]\centering
\begin{tabular}{lllrr}\toprule
Machine&Year&CPU&Cores&WPS\\\midrule
Laptop: Vaio PCG-41412L&2012&i5-2430M&2&1066\\
Desktop: iMac 27 inch&2012& i7-3770&4&3146 \\ 
Desktop&2016&i7-6700&4&6548\\
Laptop: Dell XPS 9360&2017&i7-7500U&4&3378\\
Laptop: Dell Alienware 13R3&2017&i7-7700HQ&4&5888 \\
Desktop&2019&AMD Ryzen 3600X& 6& 8791 \\ 
Desktop&2019&i7-9700& 8& 9401 \\ 
\midrule
AWS c5.metal&2019&2x8275CL&48&70037\\\bottomrule
\end{tabular}
\caption{\label{consumer}Translation speed, in words per second (WPS), of the English$\rightarrow$German model with 8-bit precision on various hardware.  Translation used all cores.  The table shows physical core count, not hyperthread count. WPS is averaged over 1M sentences.  The timing measurement includes loading time but excludes sentence splitting, which was done in advance for this experiment.}
\end{table*}

Our models are built with knowledge distillation \cite{kim-rush:2016:EMNLP2016}, use lexical shortlists \citep{schwenk-etal-2007-smooth,le-etal-2012-continuous, devlin-etal-2014-fast, Bogoychev-wngt20} to reduce the size of the output layer, 8-bit integer arithmetic, and the simplified simple recurrent unit \cite{wngt19-uedin} for decoding.  

We tested translation speed on a range of consumer hardware, shown in Table~\ref{consumer}, using the million-sentence test set from the WNGT 2020 efficiency shared task and the \textit{tiny11} preset English-German translation model from \citet{Bogoychev-wngt20}.  This test set is already sentence split, so we did not include sentence splitting and format preservation in timing.  Translating a million sentences provides ample opportunity to batch sentences of similar length and use all threads; users translating a few sentences will see slower throughput, but lower latency.


All of our student models available in the initial release are trained with the same \textit{tiny11} configuration preset. Training, knowledge distillation and quantisation instructions are described in detail on github.\footnote{\url{https://github.com/browsermt/students/tree/master/train-student}} Users can follow those instructions to train, distil and quantise their custom models, achieving noticeable speedup over vanilla \textit{float32} marian models, although any marian compatible models are supported in principle.

\subsection{Language pairs}

Our initial release includes 10 language pairs built for the Bergamot project (Table~\ref{language_pairs}). We report average BLEU scores on WMT test sets up to WMT19 \citep{wmt19}, for all languages except for Icelandic and Norwegian. For Icelandic and Norwegian we report BLEU scores on self-crawled TED Talks test set, available on github.\footnote{\url{https://github.com/browsermt/students/tree/master/isen/data} \\ \phantom{tst} \url{https://github.com/browsermt/students/tree/master/nnen/data} \\ \phantom{tst} \url{https://github.com/browsermt/students/tree/master/nben/data}}
\begin{table}
\centering
\begin{tabular}{lr}
\toprule
Languages pair & BLEU \\
\midrule
en-es & 35.0 \\
es-en & 35.3 \\
en-et & 25.1 \\
et-en & 30.8 \\
cs-en & 33.2 \\
en-cs & 25.9 \\
en-de & 41.8 \\
is-en & 23.7 \\
nn-en & 41.7 \\
nb-en & 42.7 \\
\bottomrule
\end{tabular}
\caption{Language pairs and their BLEU scores in the initial release.}
\label{language_pairs}
\end{table}

The models are distributed together with a lexical shortlist in an archive that is approximately 15MB in size. We are building and adding new models to the project.

\subsection{GUI and user interaction}

We chose the Qt\footnote{\url{https://www.qt.io}} framework to build our graphical interface. The Qt framework is widely used, open source, free for non-commercial use and in active development. We support building against both Qt5 and Qt6, which allows us to support older Linux software distributions, like Ubuntu 16.04, which do not have easy access to Qt6 packages. 

\begin{figure}[ht!]
    \centering
    \includegraphics[width=0.5\textwidth]{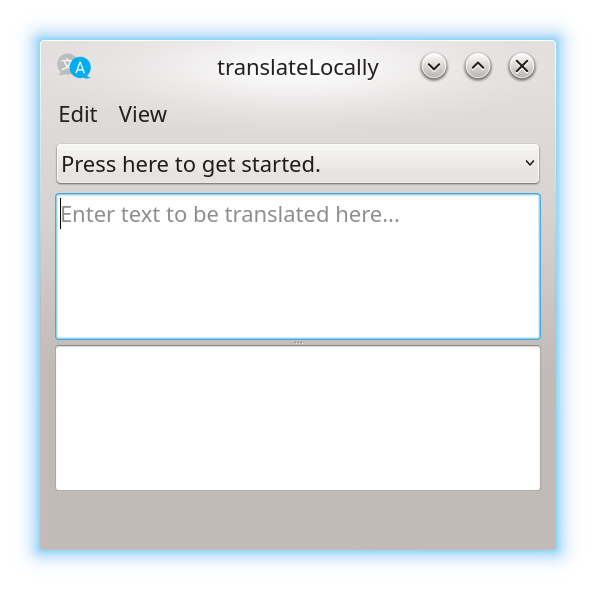}
    \caption{First run view of \textit{translateLocally}.}
    \label{fig:firstrun}
\end{figure}

We took a minimalist approach the GUI, where the user is presented with a drop-down menu to select or download models, as well as a resizeable box where the user may input text. Translations will be shown underneath or besides the input text. Translations will start to appear as soon as the user begins inputting text. The view of the first run of the program is shown in Figure~\ref{fig:firstrun}.

\begin{figure}[ht!]
    \centering
    \includegraphics[width=0.48\textwidth]{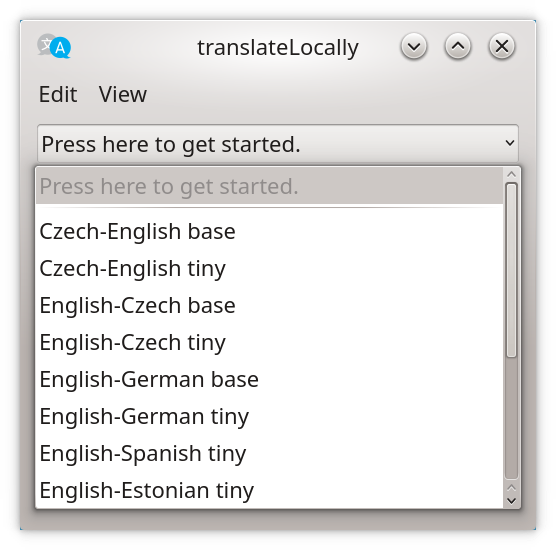}
    \caption{Select a model to download.}
    \label{fig:download}
\end{figure}

Downloading models from the Internet is done through a drop-down menu, as shown on Figure~\ref{fig:download}.  In line with our privacy promise, the application only uses Internet access following explicit user action: to retrieve the list of available models and to download a new model.  These downloads are static files.  The HTTP request includes a user-agent field with the application version number.  There is no cookie or other unique identifier.  The directory containing downloaded models can also be copied to another machine to setup a system without Internet access; we are planning to ship a version with models included.  

\begin{figure}[ht!]
    \centering
    \includegraphics[width=0.48\textwidth]{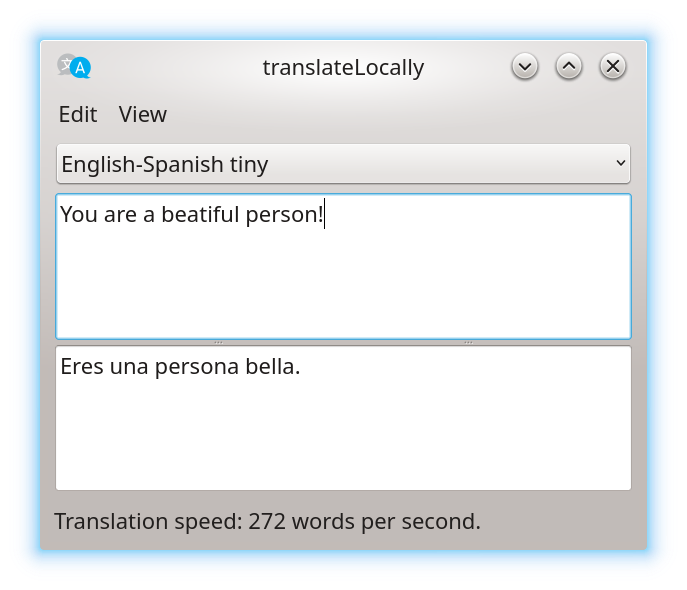}
    \caption{Translation view.}
    \label{fig:beautiful}
\end{figure}

Once the model is downloaded, typing in the input box results in a translation, shown in Figure~\ref{fig:beautiful}.

\begin{figure}[ht!]
    \centering
    \includegraphics[width=0.5\textwidth]{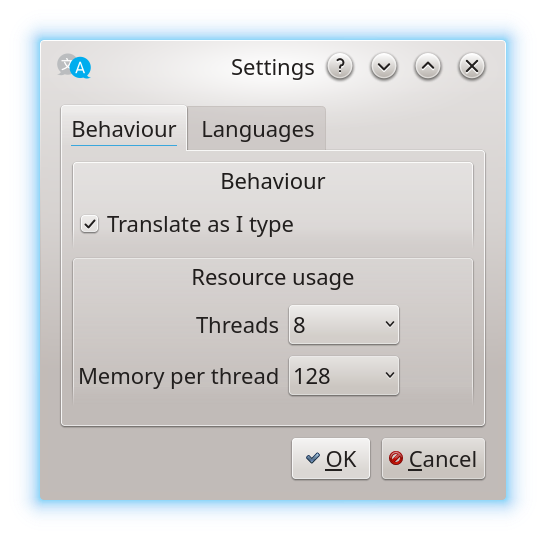}
    \caption{Settings selection for the translation engine.}
    \label{fig:settings}
\end{figure}

The application attempts to optimise thread count based on available cores and batch size based on available RAM though these can be overridden by the user, as shown on Figure~\ref{fig:settings}.  Fonts can also be changed through an OS-dependent dialog.  Retranslating as a user types consumes power, so this feature can be disabled.  

\begin{figure}[ht!]
    \centering
    \includegraphics[width=0.5\textwidth]{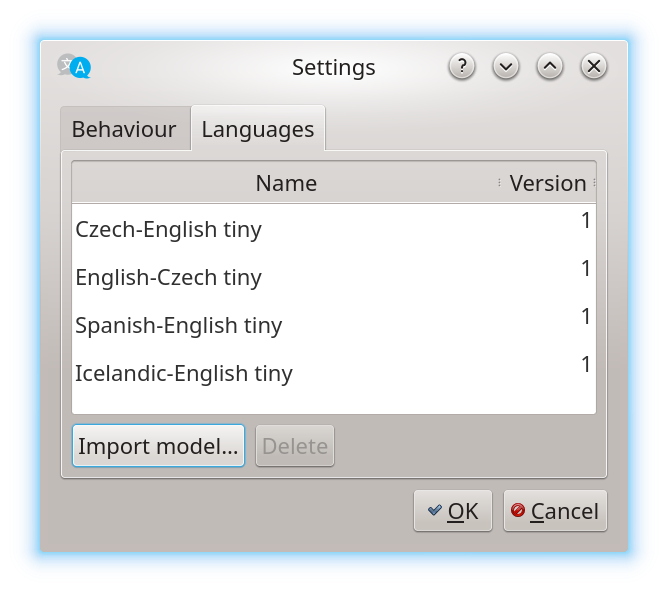}
    \caption{Model management and import window.}
    \label{fig:modelmanager}
\end{figure}

We also provide a model management screen where a user may delete downloaded models, or import custom models, as shown in Figure~\ref{fig:modelmanager}.

\begin{figure}[ht!]
    \centering
    \includegraphics[width=0.5\textwidth]{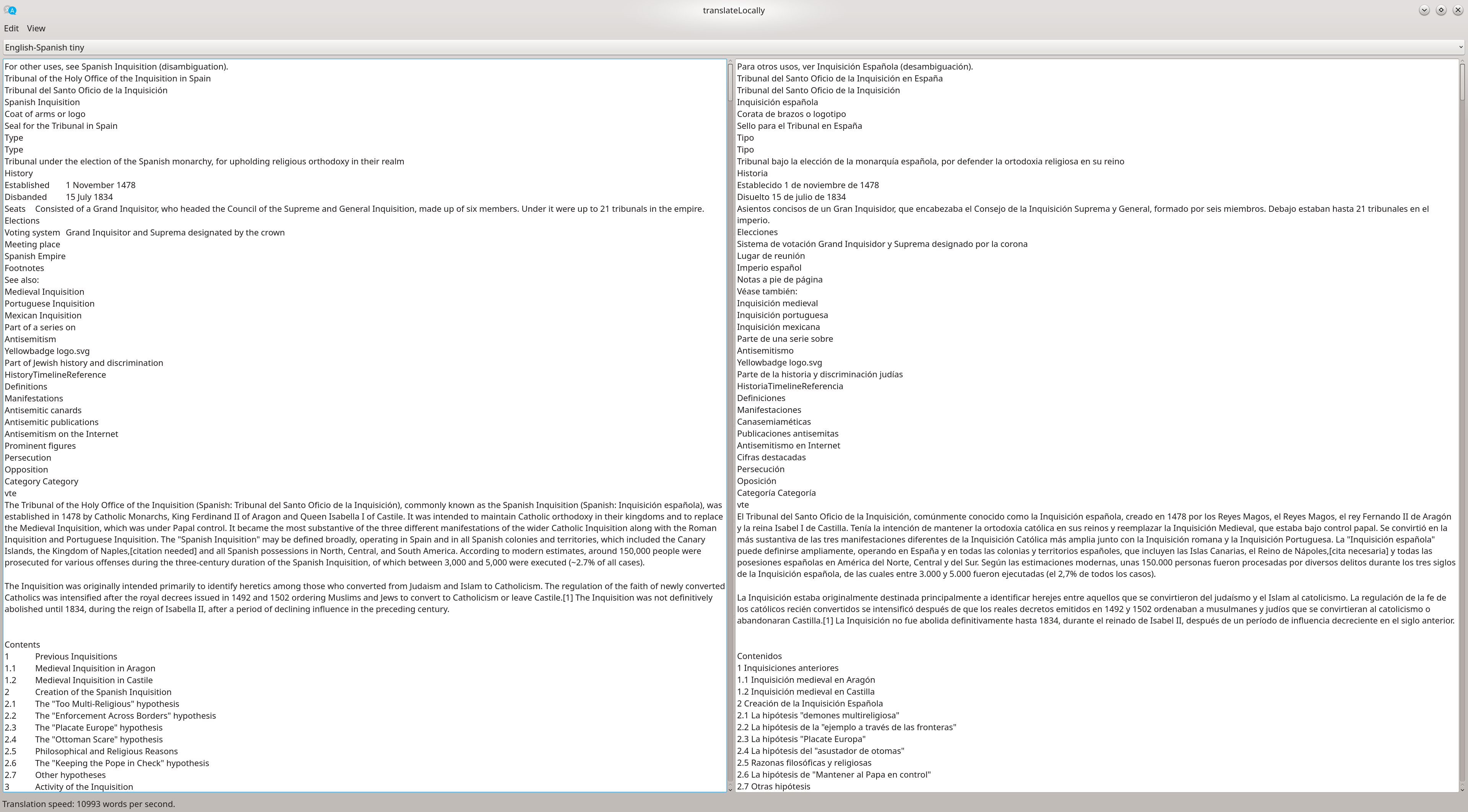}
    \caption{Translating a Big chunk of text from Wikipedia, with preservation of formatting.}
    \label{fig:spanishinq}
\end{figure}

Our translation engine preserves whitespace between sentences, so users can copy/paste content and get a well formatted text, as shown on Figure~\ref{fig:spanishinq}, which also features the side-by-side view mode.

\subsection{Distribution}
Precompiled and packaged binaries for Windows, macOS and Ubuntu 20.04 are available on the official website. Users may fetch the source code from GitHub and build it on their local machines using \textit{CMake}. The matrix multiplication library manually dispatches SSSE3, AVX2, AVX512BW and AVX512VNNI implementations based on CPUID.  However, other kernels like activation functions are currently compiled without multiple versions and will be somewhat faster if compiled explicitly for a particular vectorised instruction set.  

\section{Comparison against existing solutions}

We compare against two existing desktop machine translation solutions: \textit{Argos Translate}\footnote{\url{https://github.com/argosopentech/argos-translate}} and \textit{OPUS-CAT MT Engine} \citep{opus-cat}. They both have slightly different use-cases and support different translation languages. We compare BLEU scores \citep{bleu} on a WMT19 test set \citep{wmt19} for the English-German language pair, as well as wall-clock and CPU time. We measure only the time necessary for the actual translation. We ignore startup time and issue a translation of an unrelated text before running our test in order to discard any lazy initialisation time.

As only \textit{translateLocally} supports all three platforms, we do pairwise comparison, once on Windows for OPUS-CAT vs \textit{translateLocally}, and once on macOS for Argos Translate vs \textit{translateLocally}.

\subsection{Quality comparison}
For the quality comparison we used the following models:
\begin{itemize}
    \item For \textit{translateLocally}, we used Bergamot's English-Germany tiny model\footnote{\url{http://data.statmt.org/bergamot/models/deen/ende.student.tiny11.tar.gz}} which is just 15 MB to download.
    \item For OPUS-CAT we used the English-German opus+bt-2021-04-13.zip\footnote{\url{https://github.com/Helsinki-NLP/Tatoeba-Challenge/tree/master/models/eng-deu##opusbt-2021-04-13zip}} model, which is 275 MB in size.
    \item For Argos Translate we used their default English-German model which is downloaded through the UI, which is 87 MB in size when downloading.
\end{itemize}

We compare the BLEU scores on Table~\ref{tab:quality}.

\begin{table}[ht]
    \centering
    \begin{tabular}{lrr}
    \toprule
    System                       & Model Size & BLEU \\
    \midrule
    \textit{translateLocally}    & 15 MB  & 41.8 \\
    OPUS CAT                     & 275 MB & 40.8 \\
    Argos Translate              & 87 MB  & 34.9 \\
    \bottomrule
    \end{tabular}
    \caption{BLEU score on WMT19 English-German as well as model sizes.}
    \label{tab:quality}
\end{table}

\textit{translateLocally}'s student architecture, coupled with 8bit integer model compression delivers the smallest model size and the highest BLEU score. OPUS CAT has a comparable BLEU score, but the model is more than 15 times larger compared to \textit{translateLocally}. Argos Translate has a much lower BLEU score than either of the two, and a model size that is right in the middle.

\subsection{Argos Translate comparison}
Argos Translate is based on OpenNMT and supports 13 language pairs, with more planned in the future.

Argos Translate is not fully cross-platform as there are no windows binaries provided. The developers do advertise that it is possible for users to self-build the product on Windows. 

Finally, the macOS version is also available through the Apple app store, but it is paid,\footnote{Free macOS version is distributed through pip.} whereas \textit{translateLocally} is free.

We present our test results on Table~\ref{tab:argos}. Both system were tested on a MacBook Pro 16'' 2019, using 8 CPU threads to translate the whole WMT19 test set, which is around 40k tokens. CPU time was measured using the Activity Monitor, and Words per second (WPS) is approximately calculated. Argos Translate does not allow the CPU threads to be configured by the user, so we matched the number of threads they use in \textit{translateLocally}.

\begin{table}[ht]
    \centering
    \begin{tabular}{lrrc}
    \toprule
    System                     & WPS  & CPU Time & BLEU  \\
    \midrule
    \textit{translateLocally}  & 7350 & 40s   & 41.8 \\
    Argos Translate            & 76   & 4378s & 34.9 \\
    \bottomrule
    \end{tabular}
    \caption{\textit{translateLocally} vs Argos Translate, translating 40k tokens for speed benchmark and BLEU scores on WMT19 English-German.}
    \label{tab:argos}
\end{table}

\textit{TranslateLocally} is about 100 times faster and delivers vastly superior translation quality compared to Argos Translate.

\subsection{OPUS-CAT comparison}
Just like \textit{translateLocally}, OPUS-CAT MT Engine \citep{opus-cat} uses Marian as its translation engine. Unlike \textit{translateLocally}, its translation engine is not optimised for speed. Furthermore the GUI is slow when handling large amounts of text. Simply pasting large chunks of text, such as the full ``Crime and Punishment''\footnote{\url{https://www.gutenberg.org/files/2554/2554-0.txt}} into OPUS-CAT takes nearly as long as \textit{translateLocally} takes to paste \emph{and translate} all the text. 

The strength of OPUS-CAT comes from its plugins that integrate it with popular professional translator software, whereas our product does not support any CAT software.

OPUS-CAT has more language pairs available, which could also be used with \textit{translateLocally}, but they are not optimised for speed.

Finally OPUS-CAT is only available for Windows, as it is build using the dot NET framework, whereas \textit{translateLocally} is cross-platform.

For comparing OPUS-CAT vs \textit{translateLocally}, we used a single threaded mode for both applications, as we found no way to force OPUS-CAT to use multiple threads, whether it is through their translation interface, or through their memoQ plugin.\footnote{\url{https://www.memoq.com}} We tested on a Windows Machine with 4 CPU core i9-9800H inside Parallels, measuring the CPU time from the task manager. We pre-split the input of OPUS-CAT, as it doesn't have its own sentence splitter. Furthermore we excluded the copy/paste time from the OPUS-CAT measurements, as its XAML user interface is bad at handling large amounts of text.  We present our results on Table~\ref{tab:opuscat}.

\begin{table}[ht]
    \centering
    \begin{tabular}{lrrc}
    \toprule
    System                     & WPS  & CPU Time & BLEU  \\
    \midrule
    \textit{translateLocally}  & 1250 & 34s   & 41.8 \\
    Opus-CAT                   & 12   & 3363s & 40.8 \\
    \bottomrule
    \end{tabular}
    \caption{\textit{translateLocally} crippled to run single-threaded vs Opus-CAT, translating 40k tokens for speed benchmark and BLEU scores on WMT19 English-German.}
    \label{tab:opuscat}
\end{table}

Even with the added benefit of sentence-splitting and ignoring copy/paste time, and forcing single-threaded mode, OPUS-CAT is about 100 times slower than \textit{translateLocally}.

\section{Conclusion}

We presented \textit{translateLocally}, a desktop translation application, capable of high speed translations on a variety of hardware. Our software provides a viable alternative to cloud translation for users who are conscious of their privacy. Our product is 100 times faster than competing software and has none of the rate limitations of freemium cloud providers. We start with 10 high quality, optimised models and we aim to continuously add additional language pairs. As our product is open-source and cross-platform, it can be adopted by a wide range of users. The use of Marian as a translation engine allows for users to easily train their own models, potentially facilitating internal use for large organizations.

\section*{Acknowledgements}
We thank all the researchers and engineers working on the Bergamot\footnote{\url{https://browser.mt/partners/}} project for making \textit{translateLocally} possible, with special thanks to Ulrich Germann and Jerin Philip for their help with making the codebase cross platform, and to Graeme Nail for helping with deployment.
We thank everyone who tested the beta version of the application and the reviewers for their comments and suggestions.

\lettrine[image=true, lines=2, findent=1ex, nindent=0ex, loversize=.15]{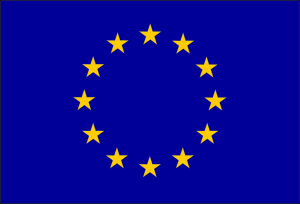}%
{T}his work was conducted within the scope of the \ProjectType\ \textit{\ProjectName}, which has received funding from the European Union's Horizon 2020 research and innovation programme under grant agreement No \GrantNo.

\bibliography{anthology,custom}
\bibliographystyle{acl_natbib}

\appendix



\end{document}